\pdfoutput=1

\documentclass[11pt]{article}

\usepackage[final]{ACL2023}
\usepackage{color,soul}
\usepackage{graphicx}
\usepackage{times}
\usepackage{latexsym}
\usepackage{amsmath}
\usepackage{amsthm}
\usepackage{mathrsfs}
\usepackage{amsfonts}
\usepackage[utf8]{inputenc}
\usepackage[sorting=none, maxbibnames=4]{biblatex} 
\usepackage{bbm}
\newtheorem{theorem}{Proposition}
\usepackage{cleveref}
\crefname{proposition}{Proposition}{propositions}
\crefname{expression}{Expression}{expressions}

\usepackage{textgreek}

\usepackage[T1]{fontenc}

\usepackage{microtype}

\usepackage{inconsolata}
\usepackage[greek.polutoniko, english]{babel}
\usepackage{blindtext}

\usepackage{csquotes}

\usepackage{fancyhdr}
\DeclareMathOperator*{\argmin}{\arg\!\min}
\DeclareMathOperator*{\argmax}{\arg\!\max}

\title{Logion: Machine Learning for Greek Philology}

\author{Charlie Cowen-Breen \
\thanks{\quad Department of Pure Mathematics and Mathematical Statistics, University of Cambridge} \
\And Creston Brooks \
\thanks{\quad Department of Computer Science, Princeton University} \
\And Johannes Haubold \
\thanks{\quad Department of Classics, Princeton University} \
\And Barbara Graziosi \
\footnotemark[3]}
\begin{document}

\maketitle
\begin{abstract}
This paper presents machine-learning methods to address various problems in Greek philology. After training a BERT model on the largest premodern Greek dataset used for this purpose to date, we identify and correct previously undetected errors made by scribes in the process of textual transmission, in what is, to our knowledge, the first successful identification of such errors via machine learning. Additionally, we demonstrate the model's capacity to fill gaps caused by material deterioration of premodern manuscripts and compare the model's performance to that of a domain expert. We find that best performance is achieved when the domain expert is provided with model suggestions for inspiration. With such human-computer collaborations in mind, we explore the model's interpretability and find that certain attention heads appear to encode select grammatical features of premodern Greek. 
\end{abstract}

\section{Introduction}\label{sec:intro}
Premodern Greek texts have been preserved in manuscripts which feature gaps caused by the deterioration of materials (e.g. papyrus, parchment, paper) and errors introduced by scribes. In the case of some premodern texts, scribes copied what survived from earlier exemplars, now lost, in a long process of textual transmission. Homer’s \textit{Iliad}, for example, dates to the 8th c. BCE and reaches us via a process of hand-copying, in relay, over many centuries: the first extant papyrus fragments date to the 3rd c. BCE and the first manuscripts of the whole poem to the 10th c. CE. To produce editions of premodern Greek texts, philologists identify scribal errors which accrued in the course of textual transmission, try to emend them, and fill gaps caused by material deterioration.

In a related study, Assael et al. train a multi-task transformer-based model to date, place, and fill gaps in ancient Greek inscriptions \cite{assael_ithaca}. Inscriptions display the original text, whereas most of what survives from antiquity reaches us via a long tradition of hand-copying from earlier exemplars. For this reason, Assael et al. focus on gaps caused by physical damage but not on copying errors.

The approaches outlined here are designed to serve philologists working on all Greek texts preserved via manuscript tradition. In particular, we introduce the intellectual work behind Logion, a framework for assisting philologists in their work.\footnote{\href{https://github.com/charliecb/Logion.git}{https://github.com/charliecb/Logion.git.}} The name means "oracle" in Greek, and we chose it to emphasize the need to interpret machine-generated results. In a proof-of-concept paper, we have already used Logion to find previously undetected errors in premodern Greek texts \cite{graziosi}. In this paper, we describe and study the approaches used to arrive at those results. More generally, we show that several tasks associated with philological research are suitable for contextual language models.

\subsection{Structure of the paper}
In \autoref{sec:intro}, we outline the training procedure for a premodern Greek BERT model \cite{BERT}, with what we believe to be the largest dataset used for this purpose to date. 

In \autoref{sec:errors}, we state the problem of detecting scribal errors and outline the approach we used in \cite{graziosi} to discover previously undetected errors in the work of Michael Psellus, an 11th-century Byzantine author. To validate the approach, we randomly generate errors that simulate those made by scribes and study the effectiveness of the model at discovering them.

In \autoref{sec:gaps}, we randomly generate artificial gaps and compare the suggestions of a domain expert to those produced by the model for filling these gaps. We find that gaps are filled with the highest accuracy when the domain expert is able to use the model's suggestions to inform his own, motivating future machine-human collaborations in Greek philology.

With such collaborations in mind, \autoref{sec:attention} presents a preliminary study of the BERT model's interpretability, analyzing specific attention heads which appear to encode select grammatical features of premodern Greek. 

In \autoref{sec:futurework}, we outline future steps to explore methodologies for error detection and, more generally, develop approaches that support the work of philologists editing premodern Greek texts.

\subsection{Training procedure}
We initially trained a BERT model on a dataset of 6.4 million words of premodern Greek, which we gratefully received from Pranaydeep Singh. This is the base model used in \cite{graziosi}. Singh et al. assembled this data from open-source databases, such as the Perseus Digital Library and the First1KGreek corpus, in the course of training the BERT model for ancient and medieval Greek presented in \cite{singh}. We subsequently assembled a much larger dataset of ca 70 million words.\footnote{Some texts were kindly provided to us by the Director of the Thesaurus Linguae Graecae to finetune our model but not to disseminate further; other texts were provided to us by colleagues who would be happy for us to share them with any interested parties. We can, therefore, share some but not all of our training data upon request: we are constrained by the license currently restricting access to the Thesaurus Linguae Graecae.} We divided this data into a 90-10 train-test split and trained the BERT model using two NVIDIA A100 GPUs for 200 epochs until validation loss stabilized. To prepare the tokenized input, we maximized the amount of punctuation-separated text in each input, up to a limit of 512 input tokens. We used a batch size of 16 and a mask ratio of 0.15.\footnote{At the task of 1-token prediction on our test set, the model achieves 84.4\% top-1 accuracy and 95.2\% top-5 accuracy, and obtains low pseudo-perplexity \cite{Markov}, 2.162. Note that these metrics are dependent on specific tokenizations and should not be compared to models with different tokenizations.}

To evaluate the impact of Singh et al.'s pre-training on modern Greek, we trained two models, one with random initialization and one initialized from Singh et al.'s Ancient Greek BERT \cite{singh}. Both times, we used Singh et al.'s tokenizer which had been created for Modern Greek sub-words, since they themselves fine-tuned a Modern Greek BERT. We find that both trainings converge to the same validation loss after a small number of epochs, indicating no discernible benefit from pre-training on Modern Greek. A future model may be more effective with a tokenizer optimized for Ancient Greek: see \autoref{sec:futurework}.

\section{Error detection}\label{sec:errors}
Premodern texts reach us via complicated processes of textual transmission, which sometimes result in scribal errors. These errors include simplifying difficult expressions, omissions, modifying ancient spellings to conform to later pronunciation, mistaking one word for another with a similar sound, shape, or function, etc. Traditionally, the only way to spot such errors has been through slow and attentive reading of the manuscripts. In this section, we show that ML can help detect and emend scribal errors. A "shortlist" of potential scribal errors can significantly improve both the speed and the accuracy of a human philologist. Logion has already helped identify errors that escaped the notice of philologists working on the same texts \cite{graziosi}.

\subsection{Related literature}

In English and other modern languages, previous work on textual error detection has typically focused on spelling and grammar checking \cite{deepspellcheck} \cite{turkish} \cite{english}, while textual errors introduced by scribes are often more complex (e.g. \autoref{fig:greektext}). For this reason, detecting scribal errors more closely aligns with \textit{out-of-distribution} detection, in which the task is to discern whether samples---in our case, words---are likely to have been generated by a given distribution---in our case, the author's body of work---or instead are out of distribution---i.e. the result of an error in transmission. \cite{llr} proposes the use of likelihood ratios to determine out-of-distribution samples, a metric which we slightly modify.\footnote{\cite{llr} use likelihood ratios to achieve state-of-the-art performance on out-of-distribution detection in genomics datasets, as well as for deep-generative models of images.} Sometimes error detection is validated by philological experts; at other times it is confirmed by manuscripts that were either sidelined or misread by previous scholars in the course of preparing the first or subsequent printed editions. 

\subsection{Approach}
We propose a two-stage approach for the discovery of textual errors. In the first stage, we fine-tune the premodern-Greek BERT model on the works of Psellus, whose texts we are investigating. Given a sequence of tokens $w_1,\hdots,w_n$, consider a single token $w_i$ and denote the surrounding context by $w_{-i} = (w_1,\hdots,w_{i-1},w_{i+1},\hdots,w_n)$. From the masked-language model (MLM) training task \cite{BERT}, the model learns the distribution
\begin{equation}\label{eqn:p}
    p(w|w_{-i})
\end{equation}
of tokens $w$ which occur in the $i$\textsuperscript{th} position of a sentence when surrounded by context $w_{-i}$. For inference on words comprised of multiple tokens, we extend $p$ to a distribution over sequences of tokens via beam search, as described in \autoref{sec:gaps}. Therefore, in what follows, when $(w_1,\hdots,w_k)$ is a sequence of words, rather than tokens, we will let $p(w|w_{-i})$ denote the corresponding distribution over words $w$ which is derived from Expression \ref{eqn:p} via beam search.

In a second step, described in the following section, we apply existing statistical theory to the learned distribution $p$ in order to determine the tokens which are most likely to contain errors.

\subsection{Shortlist generation}
Given a corpus, the goal is to generate \textbf{flags} (words which are particularly likely to be erroneous) to provide domain experts with a shortlist of potential errors. A word is flagged if it satisfies certain conditions based on the metrics we define below.

\subsubsection{Metrics}\label{subsubsec:Metrics}
We propose three metrics for flagging potential errors. Additional metrics may achieve higher accuracy at error detection in the future.\footnote{These metrics are certainly not the only ones that would lead a human philologist to consider a word suspicious, but they serve for now as a useful tool, as evidenced by \cite{graziosi}. In the future, we expect that more end-to-end methods---such as training for detecting errors directly---and regressions accounting for more metrics will outperform what is shown here.} That said, the metrics presented here have the benefit of being interpretable, as shown by \cref{prop:thm}.
\begin{enumerate}
    \item Given a word $w_i$ with context $w_{-i}$, the \textbf{chance} of word $i$ is defined as
    \[
    p(w_i|w_{-i})
    \]
    that is, the probability that the word exists in its given context, as determined by the model.
    \item The model's \textbf{confidence} at word $i$ is defined as
    \[
    \max_{\text{word } w} p(w|w_{-i})
    \]
    that is, the probability of the top suggested replacement in the given context around position $i$, as determined by the model.
    \item The \textbf{scribal distance} at token $i$ is defined as
    \[
    d\left(w_i, \quad\argmax_{\text{word } w} p(w|w_{-i})\right)
    \]
    where $d(x,y)$ denotes a modified Levenshtein distance between strings $x$ and $y$, with consideration given to mistakes that scribes are likely to make.\footnote{In particular, substitutions between the itacized vowels $\iota,\, \upsilon,\, \eta$ is penalized less than those between other combinations of vowels, based on the trivial errors listed in \cite{Reinsch}. This choice of edit distance is indeed a metric \cite{Levenshtein}.}
\end{enumerate}

\subsubsection{Rare words}
While low chance may seem to be the most intuitive indicator of errors, we find that the other two metrics are helpful for avoiding false positives. If chance were the only metric considered, genuine but rare words would be incorrectly flagged as errors.\footnote{This is because chance considers only the absolute probability of a word $w_i$ in context $w_{-i}$, instead of the relative probability when compared to plausible alternatives. Such relative probabilities are achieved by the chance-confidence ratio, which we present in the next section.} Moreover, scribal errors are sometimes graphically or phonetically similar to the correct text. Thus, by flagging low-chance portions of text which are close in sound or shape to high-confidence model suggestions, we greatly improve accuracy at detecting errors in comparison to using chance alone, as demonstrated in \autoref{subsubsec:Ablation}.

\subsubsection{Combining metrics}\label{subsubsec:likelihood}
Depending on interest, one can combine metrics in various ways to generate error flags. In what follows, we present two ranking schemes that appear to be effective at finding either real scribal errors or artificial errors introduced in order to test the effectiveness of our approach.

\begin{center}\bf
Chance-confidence ratio rankings
\end{center}
As a measure of likelihood for each word to be an error, we propose the quantity
\begin{equation}\label[expression]{eqn:chanceconfidenceratio}
    \frac{\textbf{chance}}{\textbf{confidence}}\quad\quad \bigg|\quad\quad \textbf{distance} \leq k 
\end{equation}
where the right-hand side indicates that the distributions $p(\cdot|w_{-i})$ used to compute chance and confidence should be further conditioned on the event that $d(\cdot, w_i)\leq k$, for some fixed $k$. In effect, this restricts the suggestions to those with scribal distance of no more than $k$ from $w_i$.

Formally, let
\[
\mathcal{W}_k(w_i) = \{w : d(w, w_i) \leq k\}
\]
Given a sentence $s=(w_1,\hdots,w_n)$, we will denote \cref{eqn:chanceconfidenceratio} by
\[
\rho_i(s) := \frac{p(w_i|w_{-i})}{\max_{w\in\mathcal{W}_k(w_i)} p(w|w_{-i})}
\]
and refer to it as the \textbf{chance-confidence ratio} at position $i$.

One natural motivation for the chance-confidence ratio is the following: suppose we are allowed to change only one character of a sentence and want to do so in such a way that it most resembles what a given author has written. Then, the character which we should change is exactly the one which would result in the smallest chance-confidence ratio of the affected word. This is formalized in the following proposition, which we prove in the Appendix.\footnote{For alterations of $k>1$ characters, the proposition generalizes to the corresponding statement with the assumption instead that $s'$ lies in the set of all sentences which differ from $s$ in a single word by at most $k$ characters.}

\begin{theorem}\label[proposition]{prop:thm}
{\normalfont \textbf{(Correspondence between chance-confidence ratio and relative probabilities of sentences)}} Let $p(s)$ be a joint distribution on sentences $s$ which satisfies Bayes' rule. Given a sentence $s$, suppose that
\[
s^* = \argmax_{s'\in\mathcal{W}_1(s)} p(s')
\]
Then $s^*=s$ if and only if $\rho_i(s) > 1$ for all $i$. Moreover, if $s^*\neq s$ and $i^*$ is the word index at which $s^*$ differs from $s$, then
\[
i^* = \argmin_{i} \rho_i(s)
\]
Furthermore, $s^*$ is obtained by replacing $w_{i^*}$ with the model top suggestion at $i^*$ restricted to $\mathcal{W}_1(w_{i^*})$.
\end{theorem}
In other words, the proposition states that, assuming a joint probability distribution exists,\footnote{There do not appear to be theoretical guarantees for this in the literature. In \cite{Markov}, the authors attempt to construct a joint distribution $p(s)$ directly by showing that BERT is a Markov random field, but the paper has since been retracted.} the chance-confidence ratio indicates the one-character alteration of $s$ which the model determines most likely to have been written by the author.\footnote{That said, care must be taken in concluding that $s^*$ was the original formulation of the author. Scribal errors may skew toward easier readings of the text and may thus increase $p$. This is an effect we consider further in \autoref{sec:futurework}.} This motivates ranking words by chance-confidence ratio $\rho_i(s)$ in order to detect plausible errors. In \autoref{subsec:artificial}, we artificially generate errors and find that the word with index
\[
\argmin_i \rho_i(s)
\]
indeed contains an error 90\% of the time, showing that such rankings are effective at detecting artificially generated errors (see  
\autoref{tab:artificial} and \autoref{fig:ablation}). Moreover, in 98\% of such instances, the top model suggestion at the erroneous word $w_i$,
$\argmax_{w\in W_1(w_{i})} p(w|w_{-i}),$
recovers the correct ground-truth word.

\begin{figure*}[ht]
    \centering
    \includegraphics[width=\textwidth]{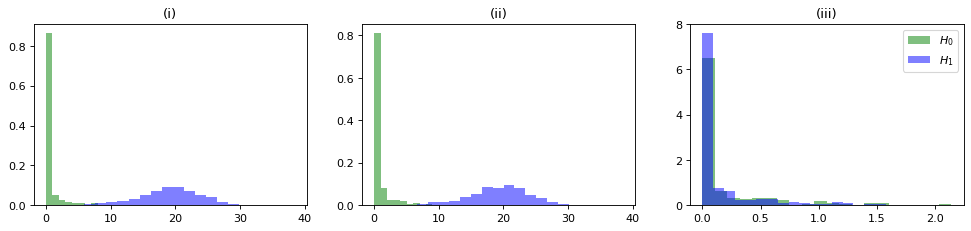}
    \caption{Distribution of metrics under hypotheses $H_0$ and $H_1$. The metrics shown are (i) chance-confidence ratio as in \cref{eqn:chanceconfidenceratio}, (ii) chance, (iii) confidence. Each horizontal scale is $-\log T$, where $T$ is the metric associated with the plot. Here, $H_1$ is modeled by the scheme to generate artificial errors described in \autoref{subsec:artificial}, where we restrict to only single token replacements in order to produce samples efficiently. Each plot contains roughly 500,000 samples from $H_0$ and 1,000 samples from $H_1$.}
    \label{fig:distribution}
\end{figure*}

Another interpretation of $\rho_i$ is that it is the likelihood-ratio statistic, assuming the prior on $w$ which is uniform on $\mathcal{W}_k$ and vanishes elsewhere. In this sense, the chance-confidence ratio builds on \cite{llr} \cite{ood}, which achieved success at detecting out-of-distribution samples with the likelihood-ratio statistic. This interpretation amounts to treating the ground truth word at position $i$ as an unknown parameter, the value of which determines the conditional distribution $p(w_{-i}|\cdot)$ of the surrounding words. In this case---again assuming that scribes only make errors which do not exceed a scribal distance of $k$---we can formulate
error detection as the hypothesis testing problem
\begin{align*}
    H_0 \quad&:\quad \text{The word $w_i$ is correct as written.} \\
    H_1 \quad&:\quad \text{The original word has been altered } \\
    &\quad\quad \text{and lies in $\mathcal{W}_k\setminus\{w_i\}$.}
\end{align*}
In \autoref{fig:distribution} (i), we plot the distribution of the likelihood-ratio statistic under the hypotheses $H_0$ and $H_1$. The distributions under each hypothesis are distinct, allowing for formal hypothesis testing via the likelihood-ratio test.

\begin{center}
\bf
Thresholding
\end{center}
In some applications, thresholding for each metric individually can provide more flexibility for generating a shortlist of errors. In \cite{graziosi}, the results were generated by thresholding for confidence of at least 50\%, scribal distance at most 3, and ranking the remaining words in order of increasing chance. In the notation of (\ref{eqn:chanceconfidenceratio}), this scheme ranks words using the quantity
\begin{equation*}
    \textbf{chance}\quad\quad\bigg|\,\,\quad \begin{array}{l}
    \textbf{distance} \leq 3 \\ \textbf{confidence} \geq 1/2 
    \end{array}
\end{equation*}
A selection of flags resulting from this scheme is shown in \autoref{subsec:novel}. The choice of a 50\% threshold for confidence is convenient because it respects the property that, among words which pass the threshold, the model's top suggestion is the same before and after thresholding for scribal distance.

Thresholds determine the precision and recall of the model when it is used to predict corrupted words. For applications where one wishes to find a list of strong candidates for erroneous words (i.e. high precision is desirable), one can set the confidence threshold to be high (e.g. 90\%) and the chance and scribal distance thresholds to be low (e.g. $10^{-6}$ and 2, respectively). For applications in which one wishes to find more corrupted words and can tolerate sifting through weaker candidates (i.e. high recall is desirable), one can set the confidence threshold to be low (e.g. 50\%) and the chance and scribal distance thresholds to be high (e.g. $10^{-4}$ and 4, respectively).

\subsection{Philologically significant results}\label{subsec:novel}
The metrics presented here have successfully identified errors that were previously undetected, ranging from scribal errors in the manuscripts, typographical errors in printed editions, and errors caused by digitization. These findings underwent philological peer review and have been accepted for publication in \textit{TAPA}, the research journal of the Society for Classical Studies.

In \cite{graziosi}, we show at proof-of-concept stage how the approaches introduced here improve on previous knowledge of premodern Greek texts by identifying and sometimes solving several different philological problems.  For detailed examples and further discussion, please see \cite{graziosi}. Below, we offer a single example to illustrate one type of error which may be detected (in this case a misreading of the manuscript rather than an actual error in the manuscript itself).

In Psellus's \textit{Hist. brev.} at lines 81.89–90, Aerts’ edition reads: 
\begin{quotation} \noindent
\textgreek{οὗτος δὶς βασιλεύσας ηὔχετο καὶ τρὶς καὶ τετράκις· ἦ δὲ γάρ, φησι, μετὰ νέφος ὁ ἥλιος.}

\noindent This man, having been king twice, prayed for a third and fourth term. \textsuperscript{\textdagger}And for indeed\textsuperscript{\textdagger}, he said, sun after clouds.
\end{quotation}

\begin{figure*}[ht]
    \centering
    \includegraphics[width=0.7\textwidth]{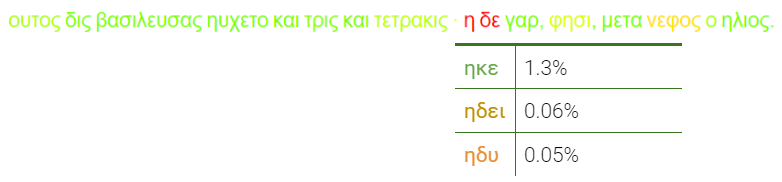}
    \caption{Algorithm output that led to the discovery of a scribal error in the words \textcolor[rgb]{1,0,0}{\textgreek{η δε}}. \textit{Top line:} words are color-coded according to their likelihood ratio, as in \autoref{eqn:chanceconfidenceratio}; the word \textcolor[rgb]{1,0,0}{\textgreek{δε}} was flagged by the algorithm because it obtained the smallest chance-confidence ratio of all words in its given context (the surrounding 512 tokens, not all of which are pictured here). \textit{Below top line:} algorithm-generated suggestions, color-coded by their chance. In each case, the algorithm suggests merging two words by deleting the space before \textcolor[rgb]{1,0,0}{\textgreek{δε}}. The third suggestion, \textcolor[rgb]{1,0.7,0}{\textgreek{ηδυ}}, is, in fact, transmitted in the relevant manuscript and must be what was originally written by the author \cite{graziosi}. The small probability awarded by Logion reflects the complexity of scribal errors. Some are trivial, including the ones we generate artificially, others, including this one, are harder to emend. 
    }
    \label{fig:greektext}
\end{figure*}

When thresholding for confidence and scribal distance, the token \textgreek{δε} was one of the lowest chance tokens in the test set. The algorithm output depicted in \autoref{fig:greektext} and the subsequent examination of the manuscript on which this edition is based led to the realization that the manuscript actually reports "\textgreek{ἡδὺ}", not "\textgreek{ἦ δὲ}". The sentence can now be translated as follows:
"This man, having been king twice, prayed for a third and fourth term. For, he said,
`sun after clouds is sweet'."

\subsection{Effectiveness}\label{subsec:artificial}
In this section, we study the effectiveness of the proposed approach for error detection by introducing artificially generated errors.
\subsubsection{Artificially generated errors}
Artificially generating scribal errors is made difficult by the fact that the data-generating mechanism is inherently complex and difficult to reproduce. Such errors are often dependent on individual scribes, the context in which they were working, and their interest in what they were copying: scribal errors can be quite varied and complex.

 That said, some errors are fairly banal, such as changes in pronunciation that can result in spelling errors due to phenomena such as itacism.\footnote{The term itacism describes a confusion between different vowels and diphthongs, all of which came to be pronounced /i/.} For the purpose of this simulation, we generate scribal errors of this kind: within every paragraph, we replace a randomly chosen character with another random character such that the modified word is in the dictionary of words used by the author at least 10 times. If the modified word does not meet this criterion, we continue substituting characters until it does. This process ensures that a simple dictionary check could not catch the errors we generate.

\subsubsection{Results}
Within every paragraph, we rank words by chance-confidence ratio (\autoref{eqn:chanceconfidenceratio}), as described in \autoref{subsubsec:likelihood} with $k=1$. Out of 615 randomly generated instances, the erroneous word ranked first 556 times, yielding a 90.5\% top-1 accuracy. Among instances in which the erroneous word ranked first, the ground-truth word was the top suggested replacement for the erroneous word 98.1\% of the time. The results are summarized in \autoref{tab:artificial}.

\begin{table*}[ht]
    \centering
    \begin{tabular}{c||c|c|c}
    \hline
        Accuracy & Chance-confidence ratio & Chance alone & Confidence alone\\
        \hline
        Top-1 & 90.5\% & 59.7\% & 54.2\%\\
        Top-5 & 95.9\% & 88.2\% & 81.1\%\\
        Top-10 & 97.6\%  & 93.1\% & 83.5\%\\
    \hline
    \end{tabular}
    \caption{Accuracy at detecting a single artificial error out of 230 words according to different schemes of combining metrics. Best performance is achieved by using chance-confidence ratio, although chance is also a viable metric. Since the task is to generate shortlists of potential errors for review (and domain experts can often verify quickly whether a flagged word is a true error) top-10 accuracy is a significant metric here.}
    \label{tab:artificial}
\end{table*}

\subsubsection{Ablation study}\label{subsubsec:Ablation}
To demonstrate that consideration of all three metrics introduced in Section 2.3 improves accuracy at detecting artificial errors, here we compare ranking by chance-confidence ratio to two alternative ranking schemes which do not involve all three metrics: (1) ranking by confidence when restricted to scribal distance 1, and (2) ranking by chance alone.

\begin{figure}[ht]
    \centering
    \includegraphics[width=0.5\textwidth]{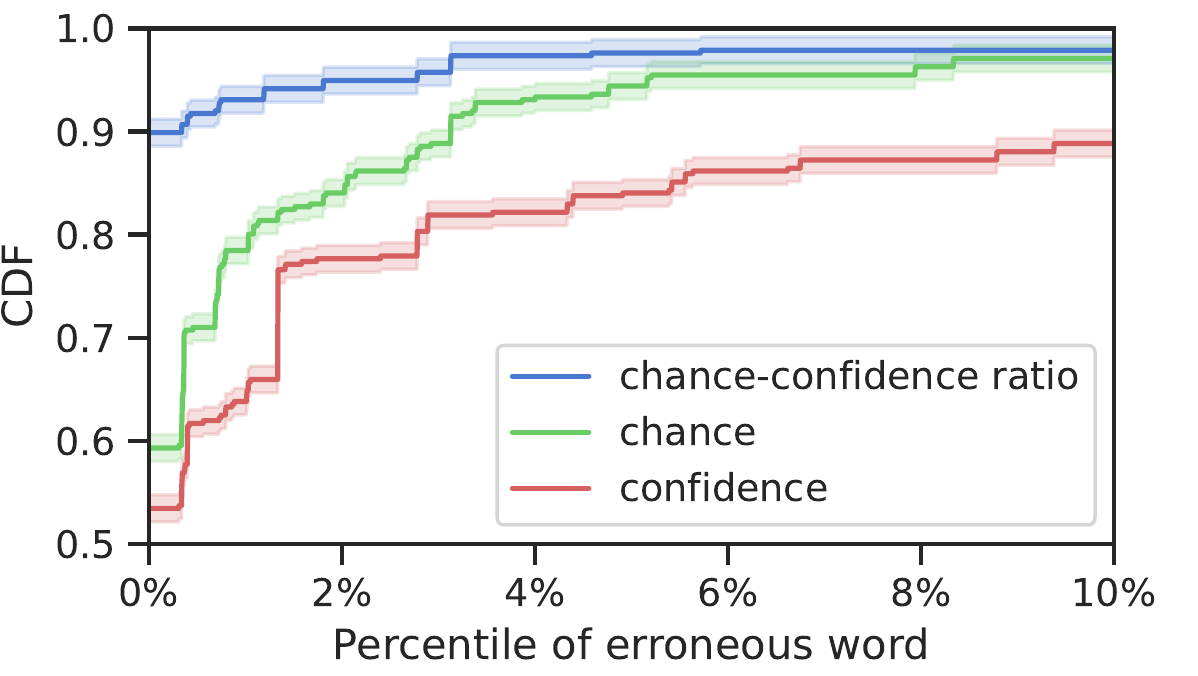}
    \caption{Artificial errors are inserted into one word per paragraph (average 230 words). The metrics of every word in the paragraph are computed (chance-confidence ratio, chance alone, confidence alone), and the percentile of the erroneous word is measured when ranked by metric within the paragraph. This plot shows the cumulative distribution function of these percentiles. 90.5\% of words rank first (i.e. 0\textsuperscript{th} percentile) in their paragraph by chance-confidence ratio, and 59.7\% of words rank first by chance, in agreement with \autoref{tab:artificial}. Error bands are computed via the DKW inequality with $99\%$ coverage probability.}
    \label{fig:ablation}
\end{figure}

\autoref{fig:ablation} shows the distribution of artificially corrupted words when ranked by the ranking schemes proposed. The distribution of each ranking scheme is heavily left-skewed: more than 85\% of erroneous words lie in the bottom 10\% of words when ranked by any metric. This suggests that each of the rankings proposed correlates with artificial errors.

However, in less than 60\% of cases is the least likely word by either of the alternative ranking schemes actually erroneous (see \autoref{tab:artificial}). Therefore, we conclude that ranking words by chance or confidence alone is less effective in identifying artificial errors than ranking by the chance-confidence ratio.

\section{Gap filling}\label{sec:gaps}
\subsection{Background}
As well as scribal errors accrued in the course of textual transmission, premodern Greek texts often suffer from gaps caused by mechanical damage or other forms of deterioration. Creating artificial gaps which resemble real ones is more straightforward than introducing artificial errors, since real gaps are effectively random portions of missing text. 

ML has already been used for the task of conjecturing what text might fit in a gap. Bamman and Burns have trained a BERT model on a large corpus of Latin texts and evaluated its accuracy for text in-filling cite{burns}. They mask single tokens, which allows them to make direct use of BERT's MLM objective but limits the usefulness of their approach to actual philological research: with real gaps, there is no way of knowing how many tokens make up a span of missing text, even if the number of missing characters can, in many cases, be estimated with some degree of accuracy. Other authors have proposed gap-filling models sensitive to gap sizes. 

For ancient Greek inscriptions, Assael et al. developed Pythia-Word, a character-level seq2seq approach utilizing both character and word representations for recovering missing characters via character-level beam search \cite{assael_pythia}. For their later model, Ithaca, input text is encoded as sequences of character, word, and position embeddings, which are passed through a transformer architecture with three task heads, specifically designed for region attribution, chronological attribution, and gap filling \cite{assael_ithaca}. Ithaca notably focuses on machine-human collaboration and sets up a mode of evaluation which we also adopt – with one modification: Assael et al. measure machine-generated results against ground truth; they then measure attempts at gap filling made by classics graduate students against ground truth; and finally, they assess results obtained via machine-human collaboration. In what follows, we adopt the same approach, except that in our case human capabilities are represented by a professional philologist who has already worked on the relevant texts using traditional methods \cite{Johannes}. We still find that human performance is enhanced by model suggestions.

\subsubsection{Model setup}\label{subsubsec:Gap filling for manuscript lacunae}
Given the number of missing tokens that make up a textual gap, an iterative beam-search heuristic over the masks can generate predictions. While the number of missing tokens in a gap is not known during inference, the number of missing characters can often be deduced from physical gap sizes; in what follows, we attempt to estimate the number of missing tokens given only the number of missing characters and the surrounding context. Consider an artificially generated gap, i.e. a gap where the ground truth is known. Let ${S_n}$ be the set of all possible spans of $n$ characters. Let \texttt{toks}($s$) denote the number of tokens that make up the span $s$, and let \texttt{gap\_toks} denote the number of tokens that make up the ground-truth text which is covered by the gap (a quantity which is unknown during inference).
\\
\begin{center}
\scalebox{0.85}{
\begin{tabular}{|c|c|}
\hline
Ground-truth & \textgreek{Οὐκ οἶδα πότερον} \\
\hline
Tokenized & [\textgreek{ουκ, οι, } -\textgreek{δα, ποτε, } -\textgreek{ρον}] \\
\hline
Gap inserted & \textgreek{Οὐκ} ???? \textgreek{πότερον} \\
\hline
\end{tabular}
}
\end{center}
\[
\texttt{gap\_toks} = 2
\]

Given context $c$ around a gap of $n$ characters, we seek $s\in S_n$ which maximizes
\begin{align*}
\mathbb{P}(s|c) =\,\,\,& \mathbb{P}(s| \texttt{gap\_toks} = \texttt{toks}(s),c)\\
\cdot\,\,\,& \mathbb{P}(\texttt{gap\_toks} = \texttt{toks}(s) | c)    
\end{align*}
We recognize $\mathbb{P}(s| \texttt{gap\_toks} = \texttt{toks}(s),c)$ as precisely $p(s|c)$, which may be computed via a beam search on $\texttt{toks(s)}$ tokens.\footnote{The true number of tokens, $\texttt{gap\_toks}$, is trivially bounded by $n$, but in practice, it is bounded by $n/2$ in more than 99\% of cases. Therefore, we perform beam searches with the number of masked tokens in the range $[1, n//2 + 2]$ to reduce computational costs.} 

To compute $\mathbb{P}(s|c)$, it now suffices to estimate $\mathbb{P}(\texttt{gap\_toks} = \texttt{toks}(s) | c)$. For this, we train a fully connected neural network (FCNN) as a function of the number of characters $n$ in the gap and the top ten probabilities from each of $(n//2 + 2)$ beam searches as input. The input dimension is $78$, created by flattening and concatenating a probability matrix of dimension $(7 \times 10)$ with a one-hot vector encoding of the number of characters. The network has one hidden layer and is trained using cross-entropy loss with $7$ output nodes representing potential values of \texttt{gap\_toks}. 

Training data is generated by randomly masking consecutive whole words in the Psellus training set until the number of masked characters, chosen randomly between 3 and 10, is exceeded. We do not count punctuation towards character count, although some gaps contain punctuation; e.g. \textgreek{ἑστηκότα· αἱ} is a 10-character gap. With each artificially masked span $s$, we generate the top ten probabilities of each beam search and use $\texttt{toks}(s)$ as the training label.

The resulting FCNN outputs a probability distribution $\mathbb{P}(\texttt{gap\_toks}=k | c)$, for $k \in [1,7]$. Thus, to make predictions for a given gap length $n$ in practice, we can perform $(n//2 + 2)$ beam searches, generate the probability distribution $\mathbb{P}(\texttt{gap\_toks}=k | c)$, and return the spans $s$ with the largest $\mathbb{P}(s|c)$ which happen to be length $n$.

\subsubsection{Consecutive masked tokens}
In the inference procedure outlined above, beam searches with up to 7 consecutive masked tokens may be performed. But inputs with many consecutive masked tokens do not align well with the BERT MLM training task: for example, with ${15\%}$ of input tokens masked, the chance of any five consecutive tokens being masked in an iteration of training is of the order $10^{-5}$.

To mitigate the discrepancy between training and inference, we fine-tune the original BERT model to handle specific lengths of consecutive masked tokens by training four new models for sequences of $2$, $3$, $4$, and $5$ masked tokens. The appropriate model is used for inferences in beam search based on the number of consecutive masked tokens in the input, with the $5$-mask model used for inputs with $\geq{5}$ consecutive masked tokens. 

\subsection{Experiments}
We conducted an experiment to compare the accuracy of our model with that of a domain expert in filling gaps in the Psellus validation set. We chose Johannes Haubold as our expert because he has already emended the text of Psellus using traditional methods \cite{Johannes}. We created 48 examples of gaps of varying lengths (3-10 characters) by randomly masking spans of consecutive whole words.\footnote{Although genuine gaps do not respect word boundaries, we masked whole words for convenience with regard to sub-word tokenization.} The number of gaps on which we tested our expert was constrained by the aim to simulate a realistic philological working environment, in which domain experts typically consider emendations over large amounts of time.

Examples contained an average of 234 words, providing sufficient context for working on the gap. The title of each example's source was also made available, as well as searchable access to the entire corpus of Greek literature, excluding the source of the example.

For each of the 48 gaps, the expert made an "expert pre-peek" prediction before being shown the model's top ten predictions and the likelihood of those predictions. He was then given the option to either stick with his original prediction or provide a new one, which we refer to as the "expert post-peek" prediction. We considered a prediction correct only if it exactly matched the ground-truth. In total, Haubold spent 7 hours filling the 48 gaps, which is considered a fast work-rate for a philologist.
 
 \subsubsection{Results and discussion}
    \begin{center}
    \begin{tabular}{c|c}
          & Accuracy\\ \hline
        Model (top-1) & 58.3\% \\
        Expert pre-peek & 64.6\% \\
        Model (top-2) & 70.8\% \\
        Expert post-peek & 75.0\% \\
        Model (top-10) & 85.4\% \\
    \end{tabular} 
    \end{center}

The study finds that the expert's initial predictions are more accurate than the model's top choice. However, when multiple suggestions are allowed, the model's accuracy greatly improves, and the expert is able to enhance his own initial conjectures by using the model's top 10 suggestions. Because the suggestions generated are intended to inspire, not replace, the philologist, we consider top-10 accuracy a good measure of success.

\section{Interpretability}\label{sec:attention}
Unlike philologists, who can supply reasoned arguments for each step of their work, the process by which BERT produces outputs is not easily understandable. In this section, we demonstrate an alignment between the grammatical observations made by scholars of premodern Greek (notably, Denniston on Greek particles \cite{denniston}) and certain attention mechanisms within the BERT architecture. Specifically, we find that certain attention heads, when interpreted as no-training-required classifiers, function surprisingly well at select grammatical tasks.

\subsection{Background and literature review}
BERT models have been found to perform remarkably well at language-based tasks, but it is not entirely clear what enables them to do so. Instead of processing linguistic information in a single, well-understood step, transformers process information through a sequence of stacked attention heads \cite{BERT}, components which are not obviously interpretable.

In \cite{englishattention}, it is proposed that some attention heads of BERT appear to perform surprisingly well when treated as no-training-required classifiers of certain grammatical tasks in English. This has motivated similar studies in several other modern languages, for example \cite{italianattention} \cite{languageattention2}.

In this section, we attempt to replicate these results for the premodern Greek BERT model.

\subsubsection{Background: attention}

For a full explanation of attention, we refer the reader to \cite{attention}. For our purposes, we note that BERT contains 12 attention heads per layer, and 12 layers, for a total of 144 attention heads.

Given some input sequence, each attention head computes a set of \textbf{attention weights} dependent on the input sequence, denoted $ \alpha_{ij}$, with one for every input token $i$ and output token $j$. Attention weights lie in the interval $[0,1]$, and attention from each token sums to 1 over the output tokens. Each attention head produces its output by summing value vectors $v_i$---linear transformations of the input---weighted by attention:
\begin{equation}\label{eqn:lincomb}
    o_i = \sum_{j=1}^n \alpha_{ij} v_j
\end{equation}
Given a specified input sequence and attention head, we will say that token $i$ \textbf{attends most} to token $j$ if
\[
j = \argmax_{j'} \alpha_{ij'}
\]
Following \cite{englishattention}, we may extend this definition to words which consist of multiple sub-word tokens in the following way: for attention from a multi-token word, we average attention weights, and for attention to a multi-token word, we sum attention weights. As the authors of \cite{englishattention} note, these transformations preserve the property that the attention from each word sums to 1.

The authors of \cite{englishattention} also observe that for specific attention heads in BERT and specific dependency relations in English, the dependency of word $i$ on word $j$ is correlated with word $i$ attending most to word $j$. Because no part of the training objective explicitly encourages attention heads to specialize in learning individual grammatical features, it is perhaps surprising that such specialization occurs.

\subsection{A case study: the \textgreek{μέν/δέ} head}
As a first case study, we focus on the construction \textgreek{μέν...δέ...}, which is conveniently made up of two indeclinable particles. In premodern Greek, these mark parallel clauses, as in the English "on the one hand..., on the other hand...". In some rare cases, \textgreek{μέν} stands alone and serves as the adverb "accordingly" or "so." Even more rarely, \textgreek{μέν} is answered by a particle other than \textgreek{δέ}, such as \textgreek{ἔπειτα} or \textgreek{μήν}.

We hypothesize that the attention head of the network at index 5-0 functions as a no-training-required classifier for the word which answers \textgreek{μέν}. If no word answers and the \textgreek{μέν} is adverbial, it attends to itself. To test this hypothesis, we first consider the distribution of words which \textgreek{μέν} attends to in the 5-0 head. In 99\% of cases, \textgreek{μέν} attends either to \textgreek{δέ}\footnote{Here we also include the elided form \textgreek{δ'}.} or to itself.

For further analysis, we distinguish between two types of cases: one in which there is a \textgreek{δέ} which appears to answer preceding \textgreek{μέν}, and one in which there is no \textgreek{δέ}. In the latter scenario, 92.2\% of the time, \textgreek{μέν} attends to itself; 5.8\% of the time, \textgreek{μέν} attends to a suitable particle other than \textgreek{δέ}, such as \textgreek{ἔπειτα} or \textgreek{μήν}.

\begin{center}
\bf
\textgreek{ἔπειτα} or \textgreek{εἶτα} as a substitute for \textgreek{δέ}
\end{center}
In cases where a corresponding clause after \textgreek{μέν} does not contain \textgreek{δέ}, it generally features \textgreek{ἔπειτα} or \textgreek{εἶτα}, and by far the most common type is when the \textgreek{μέν} clause contains \textgreek{πρῶτον}, “first" (Denniston \cite{denniston} p. 376). If the first clause contains \textgreek{πρῶτον μέν}, "first, on the one hand", the second clause is likely to contain \textgreek{ἔπειτα δέ}, "then, on the other hand", where \textgreek{δέ} becomes redundant and is easily dropped.

In alignment with these facts, \textgreek{μέν} in the premodern Greek BERT attends to \textgreek{ἔπειτα} or \textgreek{εἶτα}, in all five cases where \textgreek{πρῶτον μέν} occurs in the testing sample. \begin{center}
\bf
\textgreek{μήν} as a substitute for \textgreek{δέ}
\end{center}

In 3 of the instances when there is no \textgreek{δέ} following after \textgreek{μέν}, \textgreek{μέν} attends instead to \textgreek{μήν}, which again aligns with standard accounts of Greek particle usage. Indeed, \textgreek{μήν} is used to balance preceding \textgreek{μέν} \cite{denniston}, particularly in conjunction with the negative \textgreek{οὐ}, "not". The 5-0 attention head aligns with this grammatical observation: every time that \textgreek{μέν} attends to \textgreek{μήν}, it is when \textgreek{οὐ μήν} occurs.

\subsection{Ancient Greek dependency relations and BERT}\label{subsec:dependency}
In this section, we showcase how specific layers of the network act as classifiers for select grammatical tasks. We follow the setup in \cite{englishattention} and, in order to account for the fact that Greek is an inflected language, extract dependency relations from the Ancient Greek and Latin Dependency Treebank maintained by the Perseus Digital Library \cite{perseus}. We now test the accuracy of all attention heads at reproducing select dependency relations (\autoref{tab:grammar}), with reference to two well-studied authors of the classical period: Plato and Thucydides.

We find that while no attention head performs especially well at any particular classification of dependency relations used in \cite{perseus}, some layers appear to perform better if we focus on certain parts of speech (\autoref{tab:grammar}). As in \cite{englishattention}, we use a fixed-offset baseline as a control.\footnote{Given an attention head, we define its fixed-offset baseline in the following way: for each fixed integer $-10\leq k\leq 10$, we consider the performance of the attention head at the task of attending to the word which is $k$ positions away. The fixed-offset baseline of an attention head is defined as the maximum performance at the above task over all $k$. This baseline performance is used to test the hypothesis that, instead of truly learning grammar, given attention heads only learn to attend to nearby words.}

\begin{figure*}[ht!]
    \centering
    \includegraphics[width=0.7\textwidth]{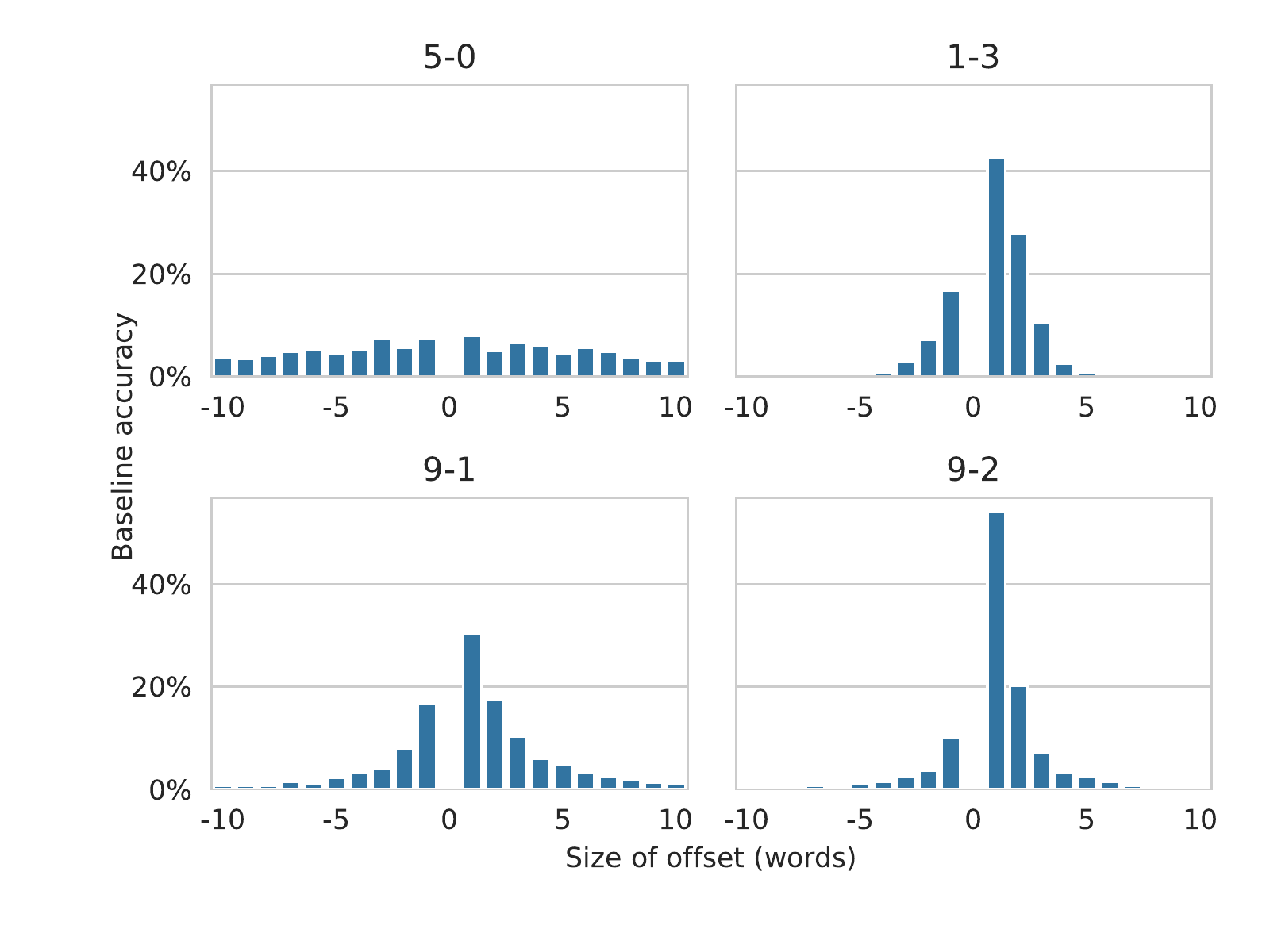}
    \caption{Comparing performance of specialized attention heads found in \autoref{tab:grammar} at baseline offset, over different sizes of offset. Heads which specialize in grammatical tasks tending to find nearby words, such as the relations between interjections and vocatives, have best performance concentrated around nearby words (i.e. small size of offset). Conversely, head 5-0, which specializes in the \textgreek{μέν/δέ} relation, has relatively low performance at any size of offset, perhaps on account of the fact that the distance between \textgreek{μέν} and an answering \textgreek{δέ} can be very large.}
    \label{fig:my_label}
\end{figure*}

As seen in \autoref{tab:grammar}, several attention heads achieve high performance at the grammatical tasks shown when compared to the fixed-offset baseline. While head 5-0 appears to specialize in the \textgreek{μέν}/\textgreek{δέ} relation, and head 1-3 appears to specialize in matching helper particles to governing verbs, heads 9-1 and 9-2 appear to specialize in general dependencies such as those found in attributive constructions. Perhaps one reason that these attention heads perform better at grammatical tasks than at the fixed-offset baseline is that word order in premodern Greek is highly flexible. For analysis of attention heads at individual fixed-offset baselines, see \autoref{fig:my_label}.

\begin{table*}
    \centering
    \begin{tabular}{c|c|c|c}
        Grammatical task & Accuracy & Specialized attention head & Fixed-offset baseline\\ \hline
        \textgreek{μέν} $\longrightarrow$ answering particle, e.g. \textgreek{δέ} & 95\% & 5-0 & 8\% (1)\\
        interjection $\longrightarrow$ vocative & 99\% & 9-1& 30\% (1)\\
    article $\longrightarrow$ articular infinitive & 90\% & 9-1& 30\% (1)\\
        corresponding particle $\longrightarrow$ optative verb  & 94\% & 1-3 & 43\% (1)\\
        
        attributive article $\longrightarrow$ noun & 86\% & 9-2& 54\% (1)\\
        attributive article $\longrightarrow$ substantive adjective & 86\% & 9-2 & 54\% (1)\\
        attributive adjective $\longrightarrow$ noun & 86\% & 9-2 &54\% (1)\\
        genitive noun in attributive position $\longrightarrow$ noun& 84\% & 9-2 &54\% (1)\\
        
    \end{tabular}
    \caption{Performance by select attention heads as no-training-required classifiers of ancient Greek dependency relations. In the right-hand column, parenthesis indicates the size of offset $k$ which maximizes performance at attending to words that are $k$ positions away.}
    \label{tab:grammar}
\end{table*}

\section{Future work}
\label{sec:futurework}
One major line of future work concerns developing an application which is adopted by domain experts and used to assist their work. Given any text, such an application would be capable of generating shortlists of suspected errors for review and of producing suggestions for how to restore missing or damaged portions of text. Future research directions in this area include developing efficient and linguistically motivated sub-word tokenization schemes, improving performance at fixed-character gap infilling, and capability to include or exclude sections of the dataset from consideration at inference time: this is relevant when one is interested in generating gap-filling suggestions which fit the work of specific authors. In working towards the latter goal, one promising architecture is DEMix \cite{demix}, which enables dynamic expert mixtures at inference time.

Another idea for future work, and one which sets scribal error detection apart from traditional error detection, concerns treating scribal modifications as diffusion processes. As scribal errors are often contextually driven, text altered by scribes may paradoxically evaluate to having higher probability than the original text.\footnote{In philology, this is the principle known as \textit{lectio difficilior potior}. Because "the normal tendency is to simplify, to trivialize, to eliminate the unfamiliar word or construction," the more difficult reading (i.e., \textit{lectio difficilior}) should normally be taken to be the authentic one \cite{west}.} On this view, then, the text evolves over time as a diffusion process with a transition kernel derived from $p$ (for example, one option is to model the trajectory of the text by Gibbs sampling according to the conditionals $p(w_i | w_{-i})$). Diffusion models \cite{diffusion} are designed to recover original data from diffused data, so it may be fruitful to apply such models for recovery of original text from scribally-modified text.

While not itself a diffusion model, ELECTRA \cite{ELECTRA} is a promising architecture for such future work. ELECTRA learns representations by simultaneously training a generator, designed to corrupt input text with plausible replacement tokens, and a discriminator, which classifies tokens in the input sequence as either replaced or original. Future work may, therefore, involve extending ELECTRA to situations in which input tokens are not randomly replaced once by the generator, but many times in a row, simulating the diffusion view of scribal error generation. By enforcing further that replaced tokens are chosen not only contextually but also according to visual and phonetic similarity, it may be possible to use such an architecture to model and detect scribal errors simultaneously.

\section{Conclusion}
\label{sec:conclusion}
In this study, we have trained a BERT model to support philological work on premodern Greek texts. We use statistical and machine-learning-based approaches to identify errors and restore gaps caused by material damage. We perform an experiment which illustrates the potential for machine-human collaboration in editing premodern texts. Finally, this paper presents a preliminary study of the BERT model's interpretability, offering insight into how specific attention heads encode select grammatical features of premodern Greek. In a broader sense, this research aims to contribute to the future of philology, understood as a discipline concerned with preserving, elucidating, and making publicly accessible the global archive of premodern texts. Some of what we have presented here is of relevance also for authors and languages we have not considered.

\section*{Acknowledgements}
We are grateful to Kasia Kobalczyk, Simon Babb, Suma Bhat, David Cox, Justin Curl, Bernhard Haubold, Max Haubold, Peter Heslin, Mika Hyman, Karthik Narasimhan, Maria Pantelia, Stratis Papaioannou, and Pranaydeep Singh for helpful conversations and advice. We are also grateful to the owners, staff, and other patrons of Bar Libreria Knulp, Trieste.


\printbibliography

\section{Appendix}
\begin{proof}[Proof of \cref{prop:thm}]
\begin{align*}
    \max_{s'\in \mathcal{W}_1(s)} p(s') &= \max_{1\leq i \leq n} \max_{w \in \mathcal{W}_1(w_i)} p(w_1,\hdots, w, \hdots w_n) \\
    &= \max_{1\leq i \leq n} \max_{w \in \mathcal{W}_1(w_i)} p(w|w_{-i}) p(w_{-i}) \\
    &= \max_{1\leq i \leq n} \max_{w \in \mathcal{W}_1(w_i)} \frac{p(w|w_{-i}) p(w_{-i})}{p(w_i|w_{-i})p(w_{-i})}p(s) \\
    &= p(s) \max_{1\leq i \leq n} \max_{w \in \mathcal{W}_1(w_i)} \frac{p(w|w_{-i})}{p(w_i|w_{-i})} \\
    &= p(s) \max_{1\leq i \leq n} \quad \frac{1}{\rho_i(s)}
\end{align*}
which establishes that, for $s^*\in\mathcal{W}_1(s)$,
\[
p(s^*) = \max_{s'\in\mathcal{W}_1(s)} p(s')
\]
if and only if $s^*$ differs from $s$ in word $i^*$ and
\[
\rho_{i^*}(s) = \min_i \rho_i(s).
\]
On the other hand, we have
\begin{align*}
    \rho_i(s) &= \frac{p(w_i|w_{-i})}{\max_{w\in\mathcal{W}_1(w_i)} p(w|w_{-i})} \\
    &= \min_{s'\in\mathcal{W}_1(s):\text{$s'$ differs from $s$ at word $i$}} \frac{p(s)}{p(s')} > 1
\end{align*}
if and only if $\forall s'$ such that $s'$ differs from $s$ only in word $i$, and only by one character, we have $p(s) > p(s')$. If this holds for all $i$, then $s=s^*$ by definition. If not, then for some $i$, it holds that $p(s) \leq p(s')$. In this case, by uniqueness of the maximum, for some $i$ for which this holds, we must have $p(s) < p(s')$. Thus $s\neq s^*$.
\end{proof}

\end{document}